\documentclass[10pt,twocolumn,letterpaper]{article}

\usepackage{cvpr}              

\usepackage[accsupp]{axessibility}  
\usepackage{graphicx}
\usepackage{amsmath}
\usepackage{amssymb}
\usepackage{booktabs}

\usepackage{array}
\usepackage{color}
\usepackage{colortbl} 
\usepackage{float}
\usepackage{subcaption}
\usepackage{multirow}
\usepackage{amssymb} 
\usepackage{makecell}
\usepackage{bm}

\usepackage[pagebackref,breaklinks,colorlinks]{hyperref}

\usepackage[capitalize]{cleveref}
\crefname{section}{Sec.}{Secs.}
\Crefname{section}{Section}{Sections}
\Crefname{table}{Table}{Tables}
\crefname{table}{Tab.}{Tabs.}

\def\cvprPaperID{4314} 
\def\confName{CVPR}
\def\confYear{2023}

\begin{document}

\title{Semantic Prompt for Few-Shot Image Recognition}

\author{First Author\\
Institution1\\
Institution1 address\\
{\tt\small firstauthor@i1.org}
\and
Second Author\\
Institution2\\
First line of institution2 address\\
{\tt\small secondauthor@i2.org}
}



\author{
Wentao Chen\textsuperscript{\rm 1,2}\footnotemark[1],
Chenyang Si\textsuperscript{\rm 3}\footnotemark[1],
Zhang Zhang\textsuperscript{\rm 2,4},
Liang Wang\textsuperscript{\rm 2,4},
Zilei Wang\textsuperscript{\rm 1},
Tieniu Tan\textsuperscript{\rm 1,2,4}\\
\textsuperscript{\rm 1} University of Science and Technology of China\\
\textsuperscript{\rm 2} Center for Research on Intelligent Perception and Computing, NLPR, CASIA\\
\textsuperscript{\rm 3} Nanyang Technological University, Singapore
\textsuperscript{\rm 4} University of Chinese Academy of Sciences\\
wentao.chen@cripac.ia.ac.cn,
chenyang.si.mail@gmail.com,
zzhang@nlpr.ia.ac.cn
}

\maketitle

\renewcommand{\thefootnote}{\fnsymbol{footnote}}
\footnotetext[1]{Equal contribution}

\begin{abstract}
   Few-shot learning is a challenging problem since only a few examples are provided to recognize a new class. Several recent studies exploit additional semantic information, \eg text embeddings of class names, to address the issue of rare samples through combining semantic prototypes with visual prototypes. However, these methods still suffer from the spurious visual features learned from the rare support samples, resulting in limited benefits.
   In this paper, we propose a novel Semantic Prompt (SP) approach for few-shot learning. Instead of the naive exploitation of semantic information for remedying classifiers, we explore leveraging semantic information as prompts to tune the visual feature extraction network adaptively. Specifically, we design two complementary mechanisms to insert semantic prompts into the feature extractor: one is to enable the interaction between semantic prompts and patch embeddings along the spatial dimension via self-attention, another is to supplement visual features with the transformed semantic prompts along the channel dimension. By combining these two mechanisms, the feature extractor presents a better ability to attend to the class-specific features and obtains more generalized image representations with merely a few support samples. Through extensive experiments on four datasets, the proposed approach achieves promising results, improving the 1-shot learning accuracy by 3.67\% on average.
\end{abstract}

\section{Introduction}

\begin{figure}
    \centering
    \includegraphics[width=0.95\linewidth]{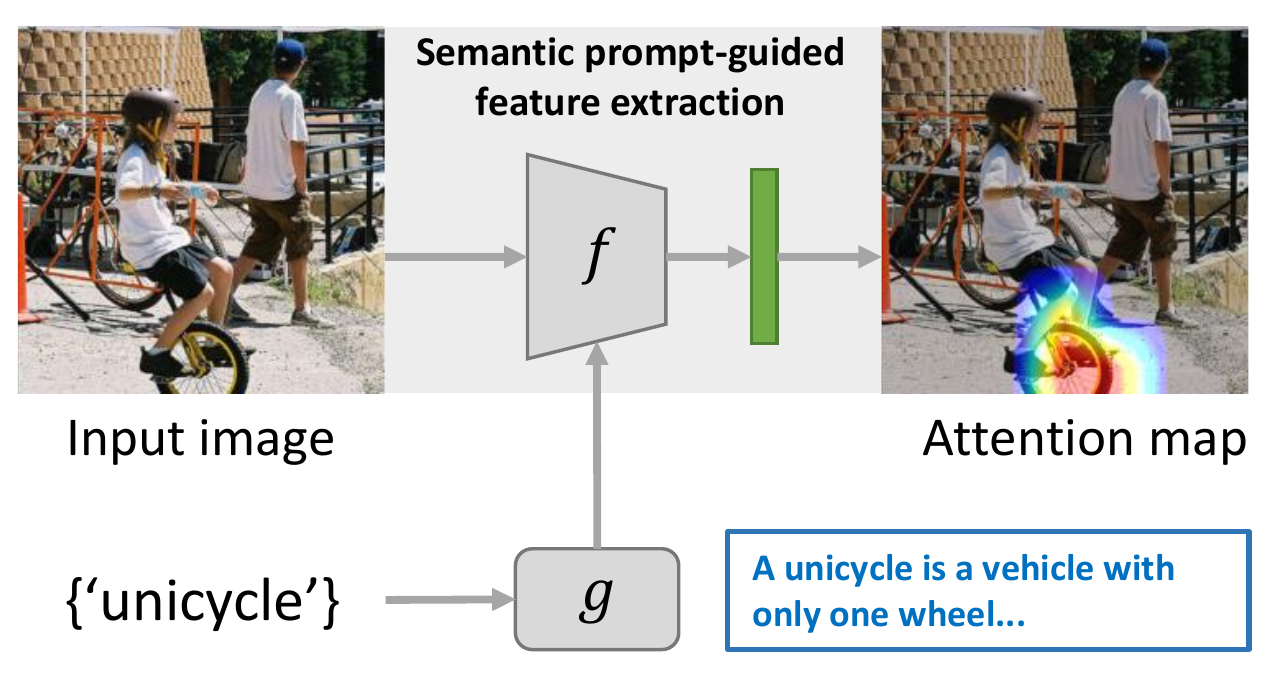}
    \caption{Given only one image about a new class `unicycle', the feature extractor is easily confused by the spurious features, such as the rider on the unicycle, and fails to obtain generalized image representations about the new class. In this paper, we propose Semantic Prompt, a new method to condition the feature extraction on rich semantic prior knowledge, such that the feature extractor captures the intrinsic class-specific features about the novel class.}
    \label{fig:introduction}
\vspace{-0.5cm}
\end{figure}

Few-shot learning (FSL) \cite{lake2011one} is a fundamental and challenging task and remains largely unsolved as it aims to predict a new class with rare samples. To address this problem, most effective FSL approaches leverage the prior knowledge learned from a large labeled base dataset, and encode the prior knowledge as a set of initial network parameters \cite{pmlr-v70-finn17a,rusu2018metalearning,Ravi2017OptimizationAA}, or a fixed embedding function shared by all classes \cite{vinyals2016matching, snell2017prototypical, sung2018learning, HUANG2021107935}.

As the labeled images of novel classes are scarce, a straightforward alternative is to use auxiliary information from other modalities, \eg natural language, to assist in learning new concepts, which has been extensively studied in zero-shot learning \cite{romera2015embarrassingly,shigeto2015ridge,li2017zero,fu2018recent}. These methods usually directly use textual embeddings as the image classifiers for novel classes. Following this idea, a recent FSL study \cite{xing2019adaptive} proposes to infer textual prototypes from class names and combine them with the visual prototypes (\ie, classifiers) extracted from the rare support images. Others \cite{peng2019few,yan2021aligning} improve this work by introducing more sophisticated textual prototype predictors (\eg Graph Convolutional Network) or producing more accurate textual prototypes through leveraging the benefits of large-scale pre-trained language models.

In spite of their success, most of the above methods for directly inferring class prototypes from textual features ignore the information gap between textual and visual features.
Specifically, the textual features may contain the semantic relationship between a novel class and known classes. However, they fail to provide the exact discriminative visual features of the new class because of lacking interaction with the underlying visual representations. As a result, the rich semantic information has derived limited benefit for recognizing novel classes when directly injecting it into classifiers.
Moreover, with only limited support images, the learned visual features still suffer from spurious features, such as background clutters, and struggles to produce an accurate class prototype. For example, as illustrated in Figure \ref{fig:introduction}, given one support image of a novel class `unicycle', the feature extractor may capture image features containing both unicycles and other distractors, like riders and tile roofs, and fail to recognize the unicycle in other environments. Actually, human perception system has a unique visual perceptual mechanism, called cognitive penetrability \cite{maier2019no}, which uses linguistic prior knowledge to tune ongoing visual perceptual processing to category-relevant stimulus features, promoting the learning of novel objects. Hence, it is necessary to develop a new architecture for effectively leveraging textual information to remedy the defective representation caused by rare samples.

In this paper, we propose Semantic Prompt, a novel approach that leverages textual information of class names to significantly improve the representation ability of visual features for few-shot learning. Instead of directly inferring prototypes from textual features, we explore leveraging the textual features as semantic prompts to adaptively tune the feature extraction network for the rare support samples. As shown in Figure \ref{fig:introduction}, with the guidance of semantic prompts, the feature extractor is expected to capture the intrinsic class-specific features for the novel class rather than other background clutters. Moreover, the advent of large-scale training has produced a cornucopia of powerful Natural Language Processing (NLP) models, such as BERT \cite{devlin2018bert} and GPT \cite{radford2018improving}, which bootstrap extracting rich textual information from class names. Through the interaction between semantic prompts and visual features, such semantically rich representations have powerful potential to provide the feature extractor with additional discriminative visual features about the new class, and subsequently produce more generalized class prototypes.

To condition the visual feature extraction on semantic prompts, we propose two complementary mechanisms to inject semantic information into the feature extractor, which allow the interaction between semantic prompts and visual features on the spatial and the channel dimensions, respectively. Specifically, to facilitate the interaction on the spatial dimension, we extend the image patch sequence with semantic prompts and feed them into a Transformer encoder. Through self-attention layers, the semantic prompts can inform the feature extractor to attend to the class-specific features while suppressing other distractors. For the interaction on the channel dimension, we first concatenate the semantic prompts with the visual context extracted from all patches, and then feed them into an MLP module. The extracted feature vector is added to each patch token to modulate and augment the visual features channel-by-channel. By combining the two interaction mechanisms, the proposed Semantic Prompt approach (SP) can effectively leverage the textual information in class names to boost FSL. Through comprehensive experiments on four benchmarks, the proposed SP presents consistent performance improvements with different types of text encoders and architecture designs, demonstrating its strong generality for the FSL problem.

In summary, our contribution are three-folds:
\begin{itemize}
    \item We propose a novel Semantic Prompt approach to leveraging textual information in class names for few-shot image recognition, which is inspired by the top-down cognitive penetrability effect in human perception and aims to adaptively tune the feature extraction to class-specific features according to the semantic prompts.
    \item To condition visual feature extraction on semantic prompts, we propose two complementary mechanisms to inject semantic prompts into the visual feature extractor, which allow the interaction on the spatial and the channel dimensions, respectively.
    \item The proposed method achieves remarkable performance on four FSL benchmarks, improving the FSL accuracy by 3.67\% on average under the challenging 1-shot setting.
\end{itemize}

\section{Related work}
\textbf{Few-shot learning.} 
FSL aims to recognize novel classes given only a few examples for each class. Previous work usually adopts a meta-learning paradigm, in which a learner is trained on a sequence of few-shot training tasks (named episodes) sampled from a large base dataset in order to rapidly adapt to unseen testing tasks. In particular, optimization-based methods \cite{pmlr-v70-finn17a,rusu2018metalearning,Ravi2017OptimizationAA} aim to learn a set of optimal initial parameters  shared by all tasks with fast adaptation ability.
Metric learning-based methods \cite{vinyals2016matching, snell2017prototypical, sung2018learning, HUANG2021107935} learn a fixed embedding function, which maps input images into a low-dimension embedding space and classifies unlabeled queries according to certain distances to the support samples, \eg, Euclidean distance\cite{snell2017prototypical}, cosine-similarity distance\cite{oreshkin2018tadam}, and Earth Mover's Distance\cite{zhang2020deepemd}.

\textbf{Few-shot learning with language.}
To leverage additional information from other modalities (especially language) to help recognize novel classes, a line of recent studies\cite{xing2019adaptive,peng2019few,li2020boosting,akyurek2022subspace} propose to integrate both visual features and auxiliary text features to represent a novel class. For example, Xing et al. \cite{xing2019adaptive} propose an adaptive fusion mechanism to combine a visual prototype with a semantic prototype obtained by the word embedding of the class label. Peng et al. \cite{peng2019few} adopt a Graph Convolutional Network \cite{zhang2019graph} as the predictor to incorporate additional knowledge from a knowledge graph. Yan et al. \cite{Yan_Zhang_Hou_Wang_Bouraoui_Jameel_Schockaert_2022} propose a word vector-guided attention mechanism to obtain label prototypes for the multi-label few-shot learning problem. Different from previous work that leverages semantic information at the level of classifiers or class prototypes, we explore the auxiliary information as a kind of semantic prompt to enhance the feature extraction for the limited support samples.

\textbf{Transformer and prompt-based learning.} Transformer is general network architecture for NLP tasks \cite{devlin2018bert,yang2019xlnet,radford2018improving,brown2020language}, and has also demonstrated great potential to deal with computer vision tasks \cite{dosovitskiy2020image,liu2021swin,zhang2022nested,si2022inception}. Specially, Dosovitskiy et al. \cite{dosovitskiy2020image} propose a simple Vision Transformer (ViT) architecture that regards image patches as a sequence and inputs them into a Transformer encoder to extract visual features. Due to the limited inductive bias in its architecture design, Transformer usually requires a lot of data to learn a new task. To address this problem, prompt-based methods \cite{petroni2019language,brown2020language} have been proposed to adapt a pre-trained language model to down-stream tasks in a data-efficient way. For example, Brown et al. \cite{brown2020language} wrap the input sentence with several hand-crafted prompt words, which inform the model of the task prior knowledge and modulate the model's behavior to the desired mode. Other studies \cite{li2021prefix,zhang2021differentiable,lester2021power} propose to replace the discrete prompt words with continuous prompt vectors that are easier to optimize than the discrete prompts. Recently, Tsimpoukelli et al. \cite{tsimpoukelli2021multimodal} propose a cross-modal prompt approach, which regards image features as the prompts for language inputs to perform multimodal few-shot learning. In this paper, we propose to regard textual features as the semantic prompts for image inputs, which can tune the ongoing visual feature extraction to class-specific features and facilitate learning novel classes with few examples. As far as we know, this is the first time to adopt semantic features as prompts to tune visual feature extractors for few-shot learning. 

\section{Problem formulation}

The FSL problem is usually defined as a $N$-way $K$-shot classification task, where a model should classify a query sample $\bm{x}^q$ from the query set $Q$ into one of $N$ classes $C_{novel}$, based on a few labeled examples ${(\bm{x}^s_i, y^s_i)}_{i=1}^{N\times K}$ from the support set $S$. Since it is very difficult to train a model from scratch with the small support set $S$, a large labeled dataset $D_{base}$ is provided to pre-train the model before performing few-shot learning. Previous work usually adopts a meta-training strategy \cite{vinyals2016matching} to split the base dataset into multiple $N$-way $K$-shot episodes. Each episode also contains a support set and a query set, mimicking the few-shot learning problem during testing. Note that the base classes $C_{base}$ do not overlap with the novel classes, \ie, $C_{base}\cap C_{novel}=\phi$. Therefore, the model is expected to acquire the ability to generalize to unseen classes after meta-training. 

\textit{Variant}: In most previous work, the image label $y$ is usually represented as a one-hot vector, \eg $y=[0,1,0,0,...]$. However, this representation erases the semantic relationships among object concepts and ignores the valuable linguistic information contained in the textual labels. In this paper, we retain text labels (\eg `cat', `dog') besides the one-hot labels in order to extract semantics from text labels. We denote $y^{text}$ as the text label to distinguish it with the one-hot label $y$.

\section{Method}
\begin{figure*}
    \centering
    \includegraphics[width=0.9\linewidth]{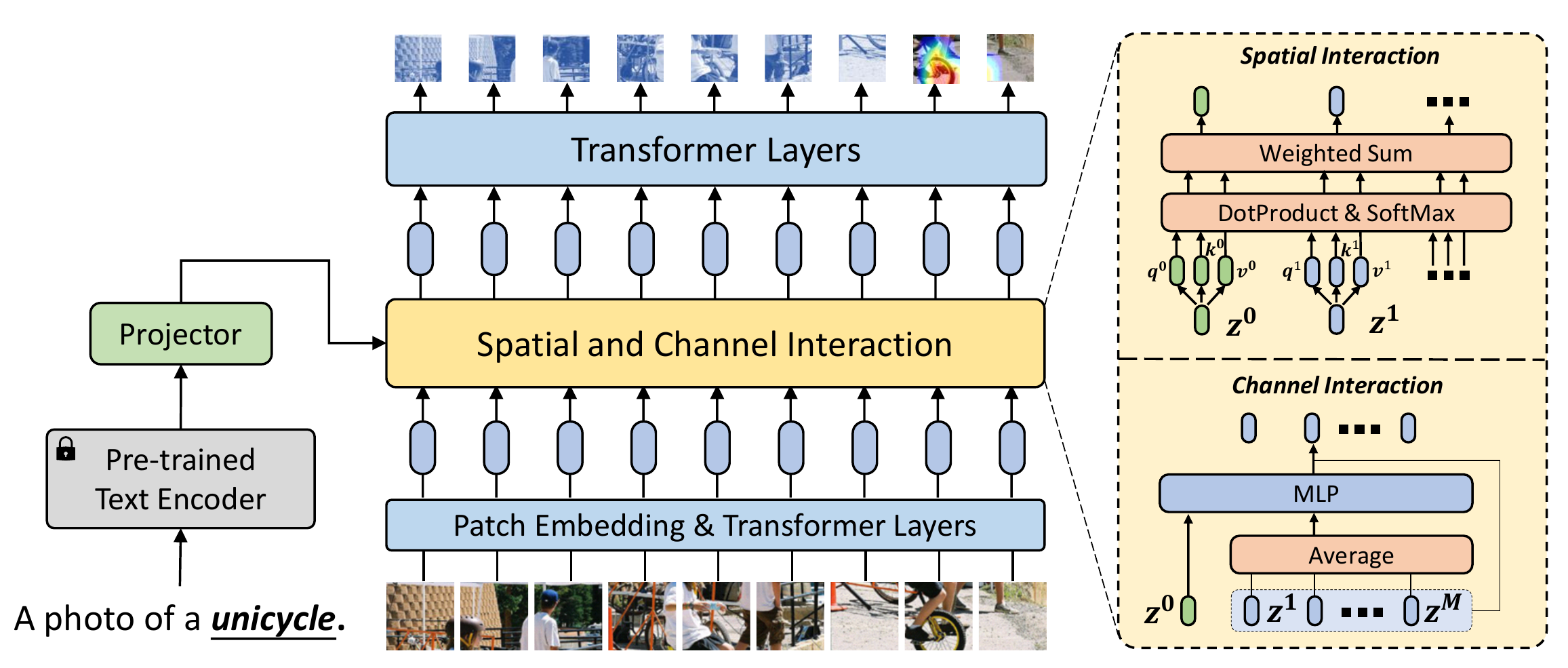}
    \caption{Framework of the proposed Semantic Prompt approach. The support image is split into small patches and fed into Transformer layers to extract visual features, which however may contain both class-specific features and other clutters. To address this problem, we leverage textual features extracted from class names as semantic prompts to adaptively tune the visual feature extraction. The semantic prompts can interact with visual features along the spatial and the channel dimensions, and guide the feature extractor to capture the intrinsic discriminative features about the new class.}
    \label{fig:framework}
\vspace{-0.5cm}
\end{figure*}

Following \cite{chen2020new}, our approach consists of two training stages. In the first stage, we pre-train a feature extractor $f$ by classifying all images in the base set $D_{base}$. In the second stage, we fine-tune $f$ with Semantic Prompt (SP) under the meta-learning paradigm, such that $f$ acquires the ability to extract generalized and class-relevant visual features for data-scarce scenarios.

\subsection{Pre-training}
\label{section:pre-train}
Learning a general feature extractor is the key to transfer knowledge to down-stream learning tasks \cite{kolesnikov2020big,he2020momentum,radford2021learning}, including few-shot learning \cite{tian2020rethinking}. Given the labeled base dataset $D_{base}$, we adopt a simple supervised learning paradigm to learn the feature extractor. A linear classification head $[\bm{W,b}]$ is added on the top of the feature extractor, which maps the input feature vector $f(\bm{x})$ into one of the base classes. We jointly train the feature extractor and the classification head by minimizing the standard cross entropy loss:
\begin{equation}
    \mathcal{L}_{pre} = \frac{1}{|D_{base}|} \sum_{(\bm{x},y)\in D_{base}} - \log \frac{\exp(\bm{W}_y^T f(\bm{x}) + \bm{b}_y)}{\sum_i \exp(\bm{W}_i^T f(\bm{x}) + \bm{b}_i)},
\end{equation}
where $\bm{W}_i, \bm{b}_i$ denote the classifier weight and the bias for the class $i$.

\textit{Backbone}: To facilitate the following interaction between visual features and semantic prompts, we adopt the Vision Transformers as the image feature extractor $f$. Specifically, an input image $\bm{x}\in \mathbb{R}^{H\times W\times C}$ is first divided to a sequence of $M$ image patches $\bm{X} = \{\bm{x}_p^1, \bm{x}_p^2,...,\bm{x}_p^M\}$ where $\bm{x}_p^i\in\mathbb{R}^{P\times P\times C}$ is an image patch and $P$ is the patch size. Then, each patch is mapped into an embedding vector and added with a learnable position embedding. The preprocessed image patches for the Transformer input can be written as: $\bm{Z}_0 = [\bm{z}_0^1, \bm{z}_0^2,...,\bm{z}_0^M]$, where $\bm{z}_0^i\in\mathbb{R}^{C_z}$ is the patch token at the position $i$ and $C_z$ is the number of channels of each token.

The patch tokens are fed into $L$ Transformer layers to extract visual features, each of which consists of multihead self-attention (MSA), a MLP block, Layernorm (LN), and residual connections. (Please refer to the appendix for more details.) At the top layer $L$, we average all embedding vectors in the sequence as the extracted image features:
\begin{equation}
    f(\bm{x}) = \frac{1}{M} \sum_{i=1}^M \bm{z}_L^i,
\label{eq:pooling}
\end{equation}
where $\bm{z}_L^i$ is the $i^{th}$ embedding vector at the layer $L$.

Note that self-attention has quadratic computation costs with respect to the sequence length. To reduce computation costs, we adopt the Visformer \cite{chen2021visformer}, a variant of the original ViT \cite{dosovitskiy2020image}, in our implementation, which replaces the first seven Transformer layers with convolutional blocks and adopts pooling among stages to reduce the sequence length.


\subsection{Semantic Prompt}


After pre-trained on the base dataset, the feature extractor $f$ can extract substantial visual features from the input images. However, due to the semantic shift between novel classes and the base dataset, the feature extractor is limited in its ability to generalize the knowledge to novel concepts with only a few labeled examples, especially when spurious correlations appear in novel class images \cite{akyurek2022subspace,wang2022model}. For example, given an image of an unseen bird standing in a tree, the model may treat both bird features and other visual features (\eg leaves, twigs) to represent the concept of the bird, and fails to recognize the bird in other environments.


To mitigate this problem, we explore additional semantic information as prompts to guide the visual feature network to obtain intrinsic and discriminative class prototypes under rare support samples, so that query images can be classified easily in terms of their distances to theses prototypes. Specifically, textual data of class names is adopted as prior knowledge for novel classes, due to its strong ability to describe semantics. Moreover, we use the NLP models with large-scale pre-training \cite{radford2021learning,reimers2019sentence,pennington2014glove} to extract textual features. The prior knowledge from a large bank of pre-trained NLP models benefits textual feature extraction from class names.

To accommodate the model to semantic prompts, we adopt the meta-training strategy \cite{vinyals2016matching} to fine-tune the feature extractor associated with semantic prompts on a series of training episodes. The framework of our approach is illustrated in Figure \ref{fig:framework}. Specifically, given a support image $\bm{x}^s$ in a training episode, we feed its class name $y^{text}$ into a pre-trained language model $g(\cdot)$ to extract semantic features \ie, $g(y^{text})$. The semantic features are used to modulate the feature extraction for the rare support samples. We denote $f_g(\bm{x}^s) = f(\bm{x}^s|g(y^{text}))$ as the conditional feature extraction process, which will be described in the following section. The obtained support features are averaged within each class to compute class prototypes. Let $\bm{p}_i$ denote the prototype for the class $i$, then
\begin{equation}
    \bm{p}_i = \frac{1}{K} \sum_{j=1}^K f_g(\bm{x}^s_j),
\end{equation}
where $\bm{x}^s_j$ is the $j^{th}$ support image of the class $i$. 

During meta-training, we freeze the text encoder $g(\cdot)$ and fine-tune other parameters by maximizing the feature similarities between query samples and their prototypes with a cross-entropy loss:
\begin{equation}
    \mathcal{L}_{meta} = - {\mathbb{E}}_{S,Q} {\mathbb{E}}_{x^q} \log \frac{\exp(s(f(\bm{x}^q), \bm{p}_{y^q})/\tau)}{\sum_{i=1}^N \exp(s(f(\bm{x}^q), \bm{p}_i)/\tau)},
\label{eq:L_meta}
\end{equation}
where $s$ denotes the cosine similarity, $\bm{p}_{y^q}$ is the prototype of the class $y^q$, and $\tau$ is a temperature hyper-parameter.

\subsubsection{Interaction on the spatial dimension}
\label{section:svp}

We first take inspiration from the prompt methods in NLP \cite{petroni2019language,brown2020language} to concatenate prompt vectors with the input sequence and feed them together into Transformer layers. Given the semantic features $g(y^{text})$ and the input sequence of patch embeddings $\bm{Z}_{l-1} = [\bm{z}^1_{l-1}, \bm{z}^2_{l-1},...,\bm{z}^M_{l-1}]\in\mathbb{R}^{M\times C_z}$ at the layer $l$, we obtain a new sequence $\hat{\bm{Z}}_{l-1}\in\mathbb{R}^{(M+1)\times C_z}$ by extending $\bm{Z}_{l-1}$ with the projected semantic features :
\begin{equation}
    \hat{\bm{Z}}_{l-1} = [\bm{z}^0, \bm{z}^1_{l-1},...,\bm{z}^M_{l-1}],
\end{equation}
where $\bm{z}^0=h_s(g(y^{text}))\in\mathbb{R}^{C_z}$ is the projected semantic embedding for spatial interaction and $h_s$ is the projector that keeps the dimension of the semantic embedding to be the same as the patch embeddings. 

Then, the extended sequence $\hat{\bm{Z}}_{l-1}$ is fed into the remaining Transformer layers, which contain multihead self-attention modules (MSA) to allow the interaction between semantic prompts and patch tokens along the spatial dimension. Specifically, letting $\hat{\bm{Z}}_{l-1}$ be the input sequence to a MSA module at the layer $l$, MSA first maps each token into three vectors, $\bm{q},\bm{k},\bm{v}\in \mathbb{R}^{N_h\times (M+1)\times C_{h}}$, with linear projection parameterized by $\bm{W}_{qkv}$, \ie,
\begin{equation}
    [\bm{q}, \bm{k}, \bm{v}] = \hat{\bm{Z}}_{l-1} \bm{W}_{qkv},
\end{equation}
where $N_h$ is the number of heads and $C_h$ is the number of channels for each head. It then computes the attention weights $\bm{A}\in \mathbb{R}^{N_h\times (M+1)\times (M+1)}$ by taking the inner product between $\bm{q}$ and $\bm{k}$ and performing softmax along the spatial dimension:
\begin{equation}
    \bm{A} = softmax(\bm{q}\bm{k}^T/C_h^{\frac{1}{4}}).
\end{equation}
The attention weights are used to choose and aggregate information from different positions. The final output is obtained by concatenating outputs of all heads and performing linear projection parameterized by $\bm{W}_{out}$:
\begin{equation}
    MSA(\hat{\bm{Z}}_{l-1}) = (\bm{A}\bm{v})\bm{W}_{out}.
\end{equation}

\subsubsection{Interaction on the channel dimension}
Besides spatial interaction via MSA, we propose another interaction mechanism that allows modulating and augmenting visual features channel-by-channel according to the input semantic prompts. Given the input sequence of patch embeddings $\bm{Z}_{l-1} = [\bm{z}^1_{l-1}, \bm{z}^2_{l-1},...,\bm{z}^M_{l-1}]\in\mathbb{R}^{M\times C_z}$ at the layer $l$, we first obtain a global visual context vector $\bm{z}^c_{l-1}\in\mathbb{R}^{C_z}$ by averaging all patch tokens:
\begin{equation}
    \bm{z}^c_{l-1} = \frac{1}{M} \sum_{i=1}^M \bm{z}^i_{l-1}.
\end{equation}

The visual context $\bm{z}^c_{l-1}$ is then concatenated with the projected semantic vector $\bm{z}^0 = h_c(g(y_{text}))\in\mathbb{R}^{C_z}$, and fed into a 2-layer MLP module to obtain a modulating vector $\bm{\beta}_{l-1}\in\mathbb{R}^{C_z}$:
\begin{equation}
    \bm{\beta}_{l-1} = \sigma(\bm{W}_2\ \sigma(\bm{W}_1\ [\bm{z}^0;\bm{z}^c_{l-1}]+\bm{b}_1)+\bm{b}_2),
\end{equation}
where $\bm{W}_1,\bm{b}_1,\bm{W}_2,\bm{b}_2$ are the parameters of the MLP module, $\sigma$ is the sigmoid activation function, and $h_c$ is the projector for the channel interaction.

We finally add the modulating vector to all patch tokens such that it can tune the visual features at each channel. The modulated sequence $\tilde{\bm{Z}}_{l-1}\in\mathbb{R}^{M\times C_z}$ can be written as:
\begin{equation}
    \tilde{\bm{Z}}_{l-1} = [\bm{z}^i_{l-1}+\bm{\beta}_{l-1},]\quad i=1,2,...,M.
\end{equation}

\section{Experiments}

\subsection{Datasets and implementation details}
\label{section:dataset}

\begin{table*}[t]
\small
\centering
    \begin{tabular}{lclcccc}
    \toprule
     & & & \multicolumn{2}{c}{\emph{mini}ImageNet 5-way} &  \multicolumn{2}{c}{\emph{tiered}ImageNet 5-way} \\
     Method  & Backbone & Params/FLOPS & 1-shot & 5-shot & 1-shot & 5-shot \\
    \midrule
     LEO \cite{rusu2018metalearning} & WRN-28-10 & 36.5M/$3.7\times10^{10}$ & 61.76$\pm$0.08 & 77.59$\pm$0.12 & 66.33$\pm$0.05 & 81.44$\pm$0.09 \\
     CC+rot \cite{gidaris2019boosting} & WRN-28-10 & 36.5M/$3.7\times10^{10}$ & 62.93$\pm$0.45 & 79.87$\pm$0.33 & 70.53$\pm$0.51 & 84.98$\pm$0.36 \\
     Align \cite{Afrasiyabi2020AssociativeAF} & WRN-28-10 & 36.5M/$3.7\times10^{10}$ & 65.92$\pm$0.60 & 82.85$\pm$0.55 & \textbf{74.40$\pm$0.68} & 86.61$\pm$0.59 \\
     MetaOptNet \cite{Lee2019MetaLearningWD} & ResNet-12 & 12.5M/$3.5\times10^9$ & 62.64$\pm$0.61 & 78.63$\pm$0.46 & 65.99$\pm$0.72 & 81.56$\pm$0.53 \\
     Meta-Baseline \cite{chen2020new} & ResNet-12 & 12.5M/$3.5\times10^9$ & 63.17$\pm$0.23 & 79.26$\pm$0.17 & 68.62$\pm$0.27 & 83.74$\pm$0.18\\
     DeepEMD \cite{zhang2020deepemd} & ResNet-12 & 12.5M/$3.5\times10^9$ & 65.91$\pm$0.82 & 82.41$\pm$0.56 & 71.16$\pm$0.87 & 86.03$\pm$0.58 \\
     RE-Net \cite{kang2021relational} & ResNet-12 & 12.5M/$3.5\times10^9$ & 67.60$\pm$0.44 & 82.58$\pm$0.30 & 71.61$\pm$0.51 & 85.28$\pm$0.35 \\
     TPMM \cite{wu2021task} & ResNet-12 & 12.5M/$3.5\times10^9$ & 67.64$\pm$0.63 & \textbf{83.44$\pm$0.43} & 72.24$\pm$0.70 & 86.55$\pm$0.63\\
     SetFeat \cite{afrasiyabi2022matching} & ResNet-12 & 12.5M/$3.5\times10^9$ & \textbf{68.32$\pm$0.62} & 82.71$\pm$0.46 & 73.63$\pm$0.88 & \textbf{87.59$\pm$0.57} \\
     SUN \cite{dong2022self} & Visformer-S & 12.4M/$1.7\times10^8$ & 67.80$\pm$0.45 & 83.25$\pm$0.30 & 72.99$\pm$0.50 & 86.74$\pm$0.33 \\
     \midrule
     KTN \cite{peng2019few} & ResNet-12 & 12.5M/$3.5\times10^9$ & 61.42$\pm$0.72 & 74.16$\pm$0.56 & - & - \\
     AM3 \cite{xing2019adaptive} & ResNet-12 & 12.5M/$3.5\times10^9$ & 65.30$\pm$0.49 & 78.10$\pm$0.36 & 69.08$\pm$0.47 & 82.58$\pm$0.31 \\
     TRAML \cite{li2020boosting} & ResNet-12 & 12.5M/$3.5\times10^9$ & \textbf{67.10$\pm$0.52} & 79.54$\pm$0.60 & - & - \\
     DeepEMD-BERT \cite{yan2021aligning} & ResNet-12 & 12.5M/$3.5\times10^9$ & 67.03$\pm$0.79 & \textbf{83.68$\pm$0.65} & \textbf{73.76$\pm$0.72} & \textbf{87.51$\pm$0.75}\\
     \midrule
     Pre-train (Ours) & Visformer-T & 10.0M/$1.3\times10^9$ & 65.16$\pm$0.44 & 81.22$\pm$0.32 & 72.38$\pm$0.50 & 86.74$\pm$0.34 \\
     \textbf{SP-CLIP} (Ours)  & Visformer-T & 10.0M/$1.3\times10^9$ & \textbf{72.31$\pm$0.40} & 83.42$\pm$0.30 &  \textbf{78.03$\pm$0.46} & 88.55$\pm$0.32 \\
     \textbf{SP-SBERT} (Ours)  & Visformer-T & 10.0M/$1.3\times10^9$ & 70.70$\pm$0.42 & \textbf{83.55$\pm$0.30} & 73.31$\pm$0.50 & 88.56$\pm$0.32 \\
     \textbf{SP-GloVe} (Ours)  & Visformer-T & 10.0M/$1.3\times10^9$ & 70.81$\pm$0.42 & 83.31$\pm$0.30 & 74.68$\pm$0.50 & \textbf{88.64$\pm$0.31} \\
    \bottomrule
    \end{tabular}
\caption{Comparison with previous work on \textit{mini}ImageNet and \textit{tiered}ImageNet. Methods in the top rows do not use semantic information, and methods in the middle rows leverage semantic information from class names \cite{peng2019few,xing2019adaptive,li2020boosting} or descriptions \cite{yan2021aligning}. Accuracies are reported with 95\% confidence intervals.}
\label{tab:mini}
\vspace{-0.4cm}
\end{table*}

\textbf{\textit{mini}ImageNet and \textit{tiered}ImageNet.} The \textit{mini}ImageNet dataset is proposed in \cite{vinyals2016matching} to benchmark the few-shot learning problem. It contains a subset of 100 classes in the ImageNet \cite{ILSVRC15} dataset, where 64 classes are used as base classes for pre-training and meta-training, 16 classes are used for validation, and 20 classes are used for testing. The \textit{tired}ImageNet 
 dataset \cite{ren2018metalearning} is also derived from ImageNet and contains more classes: 351 classes used for training, 97 classes used for validation, and 160 classes used for testing. The semantic difference between base classes and novel classes in the \textit{tiered}ImageNet dataset is much larger than \textit{mini}ImageNet.

\textbf{CIFAR-FS and FC100.} These two datasets are derived from the CIFAR-100 \cite{cifar} dataset with different partition modes. CIFAR-FS \cite{Lee2019MetaLearningWD} randomly splits 100 classes into 64 training classes, 16 validation classes and 20 testing classes. In contrast, FC100 \cite{oreshkin2018tadam} divides classes based on their semantic superclasses, where 60 classes from 20 superclasses are used for training, 20 classes from 4 superclasses are used for validation, 20 classes from 4 superclasses are used for testing. The large semantic gap makes FC100 more difficult than CIFAR-FS.

\textbf{Text encoders.} To extract rich semantic features from class names, we adopt three types of text encoders, \ie, CLIP\cite{radford2021learning}, SBERT\cite{reimers2019sentence}, and GloVe\cite{pennington2014glove}, which are pre-trained on large-scale corpora and are available for public use. For CLIP, we only use its text encoder, and extend the input class name with a text template: \textit{A photo of a \{class name\}}. For SBERT and Glove, we directly feed class names into their encoders and average the output word vectors if there are multiple words in a name.

\textbf{Implementation details.} 
We adopt Visformer-Tiny \cite{chen2021visformer} as the feature extractor and resize the input image into 224$\times$224 by default. Other input resolutions are validated in Section \ref{sec:image_size}. Images are augmented with RandomResizedCrop, RandAug \cite{cubuk2020randaugment} and RepeatAug \cite{berman2019multigrain}. During pre-training, we use the AdamW optimer \cite{loshchilov2017decoupled} with a learning rate of 5e-4 and a weight decay of 5e-2. We pre-train the model for 800 epochs on \textit{mini}ImageNet, CIFAR-FS and FC100, and for 300 epochs on \textit{tiered}ImageNet. During meta-training, we reduce the learning rate of the feature extractor to 1e-6 and set the learning rate of the projectors as 5e-4. The model is meta-trained for 100 epochs on all datasets. The hyper-parameter $\tau$ is set as 0.2 according to validation accuracy. We conduct experiments with a TITAN Xp server and training can be done with one GPU.

During evaluation, we randomly sample 2,000 test episodes from the novel classes. For 1-shot learning, we use the cosine classifier for prediction as in Eq.\ref{eq:L_meta}. For 5-shot learning, we adopt logistic regression classifiers with random crop augmentation. We finally report the average accuracy with 95\% confidence intervals.

\subsection{Comparison with the state-of-the-art}
\label{section:comparison}

\begin{table*}[t]
\small
\centering
    \begin{tabular}{lclcccc}
    \toprule
     & & & \multicolumn{2}{c}{CIFAR-FS 5-way} &  \multicolumn{2}{c}{FC100 5-way} \\
     Method  & Backbone & Params/FLOPs & 1-shot & 5-shot & 1-shot & 5-shot \\
    \midrule
     PN+rot \cite{gidaris2019boosting} & WRN-28-10 & 36.5M/$3.7\times10^{10}$ & 69.55$\pm$0.34 & 82.34$\pm$0.24 & - & - \\
     Align \cite{Afrasiyabi2020AssociativeAF} & WRN-28-10 & 36.5M/$3.7\times10^{10}$ & - & - & \textbf{45.83$\pm$0.48} & 59.74$\pm$0.56 \\
     ProtoNet \cite{snell2017prototypical} & ResNet-12 & 12.5M/$3.5\times10^9$ & 72.2$\pm$0.7 & 83.5$\pm$0.5 & 37.5$\pm$0.6 & 52.5$\pm$0.6 \\
     MetaOptNet \cite{Lee2019MetaLearningWD} & ResNet-12 & 12.5M/$3.5\times10^9$ & 72.6$\pm$0.7 & 84.3$\pm$0.5 & 41.1$\pm$0.6 & 55.5$\pm$0.6 \\
     MABAS \cite{Kim2020ModelAgnosticBS} & ResNet-12 & 12.5M/$3.5\times10^9$ & 73.51$\pm$0.92 & 85.49$\pm$0.68 & 42.31$\pm$0.75 & 57.56$\pm$0.78 \\
     Distill \cite{tian2020rethinking} & ResNet-12 & 12.5M/$3.5\times10^9$ & 73.9$\pm$0.8 & 86.9$\pm$0.5 & 44.6$\pm$0.7 & \textbf{60.9$\pm$0.6} \\
     RE-Net \cite{kang2021relational} & ResNet-12 & 12.5M/$3.5\times10^9$ & 74.51$\pm$0.46 & 86.60$\pm$0.32 & - & - \\
     infoPatch \cite{liu2021learning} & ResNet-12 & 12.5M/$3.5\times10^9$ & - & - & 43.8$\pm$0.4 & 58.0$\pm$0.4 \\
     SUN \cite{dong2022self} & Visformer-S & 12.4M/$1.7\times10^8$ & \textbf{78.37$\pm$0.46} & \textbf{88.84$\pm$0.32} & - & - \\
     \midrule
     Pre-train (Ours) & Visformer-T & 10.0M/$1.3\times10^9$ & 71.99$\pm$0.47 & 85.98$\pm$0.34 & 43.77$\pm$0.39 & 59.48$\pm$0.39 \\
     \textbf{SP-CLIP} (Ours)  & Visformer-T & 10.0M/$1.3\times10^9$ & \textbf{82.18$\pm$0.40} & 88.24$\pm$0.32 & \textbf{48.53$\pm$0.38} & \textbf{61.55$\pm$0.41} \\
     \textbf{SP-SBERT} (Ours)  & Visformer-T & 10.0M/$1.3\times10^9$ & 81.32$\pm$0.40 & 88.31$\pm$0.32 & 47.03$\pm$0.40 & 61.03$\pm$0.40 \\
     \textbf{SP-GloVe} (Ours)  & Visformer-T & 10.0M/$1.3\times10^9$ & 81.62$\pm$0.41 & \textbf{88.32$\pm$0.32} & 46.69$\pm$0.41 & 61.18$\pm$0.41 \\
    \bottomrule
    \end{tabular}
\caption{Comparison with previous work on CIFAR-FS \cite{Lee2019MetaLearningWD} and FC100 \cite{oreshkin2018tadam}.}
\label{tab:cifar}
\vspace{-0.3cm}
\end{table*}

To evaluate the effectiveness of our approach, we conduct extensive experiments on four datasets
, and compare the results with previous state-of-the-art methods in Table \ref{tab:mini} and Table \ref{tab:cifar}.

Compared with previous methods that leverages semantic information (KTN \cite{peng2019few}, AM3 \cite{xing2019adaptive}, TRAML \cite{li2020boosting}, Deep-BERT \cite{yan2021aligning}), our method improves 1-shot accuracy by 5.21\% on \textit{mini}ImageNet and by 4.27\% on \textit{tiered}ImageNet. DeepEMD-BERT achieves better 5-shot accuracy than ours on \textit{mini}ImageNet, but requires multiple forward passes and additional inner optimization step to obtain reliable local feature similarities. Note that previous methods usually adopts CNN as the backbone, except a recently proposed method SUN \cite{dong2022self} that also adopts the Visformer backbone. Nevertheless, our method outperforms SUN by 2.46\% on average over three datasets.

When using different text encoders to extract semantic features, the proposed SP presents consistent improvements over the pre-training baseline. Specifically, we can see that SP with CLIP achieves better on 1-shot than SBERT and GloVe, probably because CLIP's multi-modal pre-training results in better alignment of semantic embeddings with visual concepts. In 5-shot, the performance difference decreases as the model performance is dominated by visual features when support images are sufficient. In the following experiments, we use CLIP as the default text encoder.

\subsection{Model analysis}

\subsubsection{Ablation study}
\label{section:ablation}

\begin{table}[t]
\small
\centering
    \begin{tabular}{ccccccc}
    \toprule
    Aug & SI & CI & Mini & Tiered & CIFAR-FS & FC100 \\
    \midrule
    $\times$ & $\times$ & $\times$ & 61.96 & 71.91 & 68.84 & 40.78 \\
    $\checkmark $ & $\times$ & $\times$ & 65.15 & 72.38 & 71.99 & 43.77 \\
    $\checkmark $ & $\checkmark$ & $\times$ & 71.59 & 76.20 & 81.19 & 47.83 \\
    $\checkmark $ & $\times$ & $\checkmark$ & 70.48 & 77.62 & 79.80 & 47.10 \\
    $\checkmark $ & $\checkmark$ & $\checkmark$ & \textbf{72.31} & \textbf{78.03} & \textbf{82.18} & \textbf{48.53} \\
    \bottomrule
    \end{tabular}
\caption{Ablation study on four datasets under the 1-shot setting. SI means spatial interaction, and CI means channel interaction.}
\label{tab:ablation}
\vspace{-0.5cm}
\end{table}

The ablation study results are shown in Table \ref{tab:ablation}. By extending the standard RandomResizedCrop with RandAug and RepeatAug, the 1-shot accuracy of the pre-trained feature extractor is improved by 2.45\% on average over four datasets. To validate the effectiveness of SP, we fine-tune the feature extractor with three different interaction mechanisms, including SI (spatial interaction), CI (channel interaction) and SI+CI. As shown in Table \ref{tab:ablation}, both SI and CI are very effective, improving average 1-shot accuracy on 4 datasets by 5.89\% and 5.43\%, respectively. Furthermore, by combing them together, the 1-shot learning accuracy is further improved on all four datasets. These results indicate that the proposed SP is an effective approach to leveraging semantic information for few-shot learning.

\vspace{-0.1cm}
\subsubsection{Layer selection}
\label{section:layer}

\begin{figure}
    \centering
    \subcaptionbox{}{\includegraphics[width=.49\linewidth]{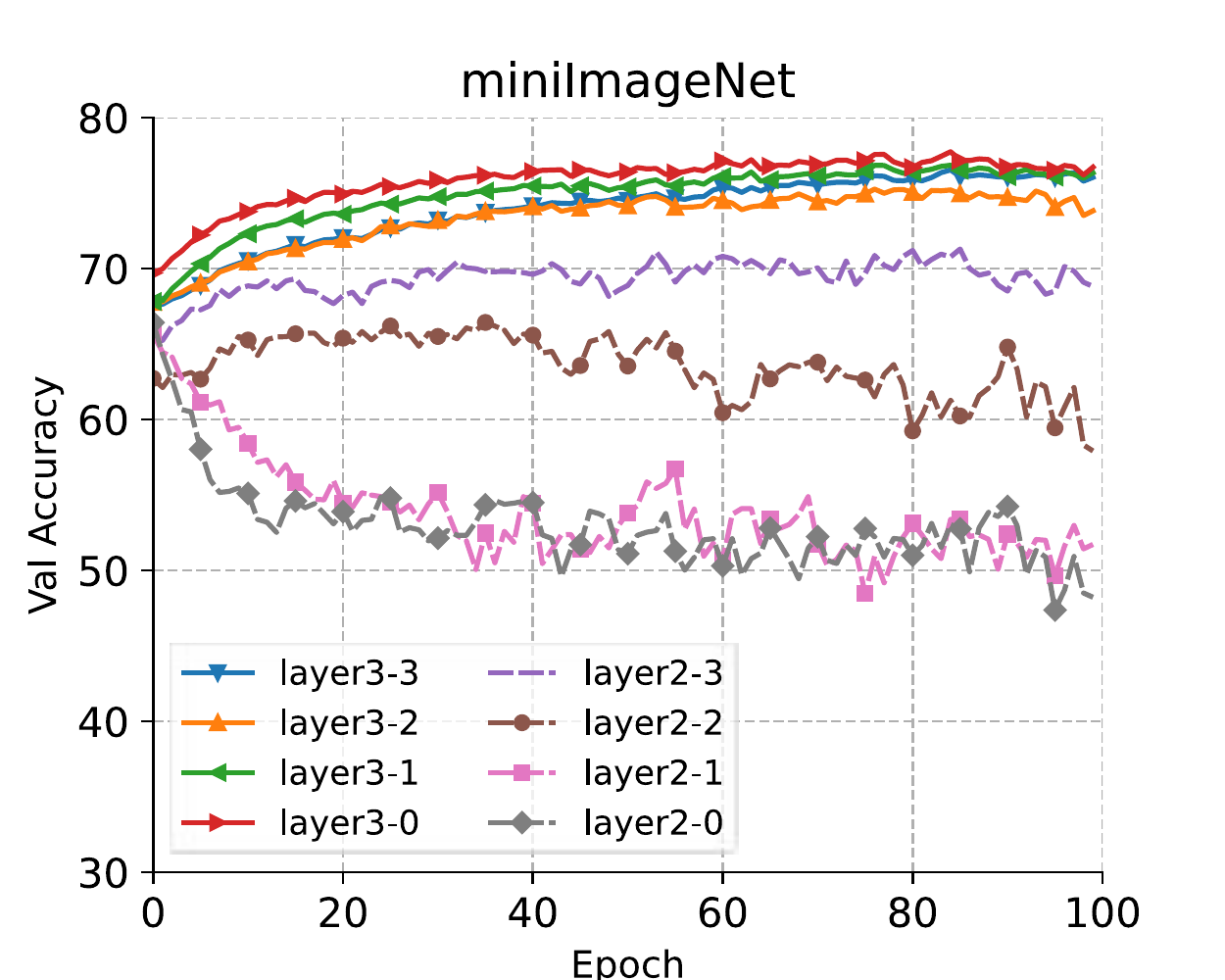}}
    \subcaptionbox{}{\includegraphics[width=.49\linewidth]{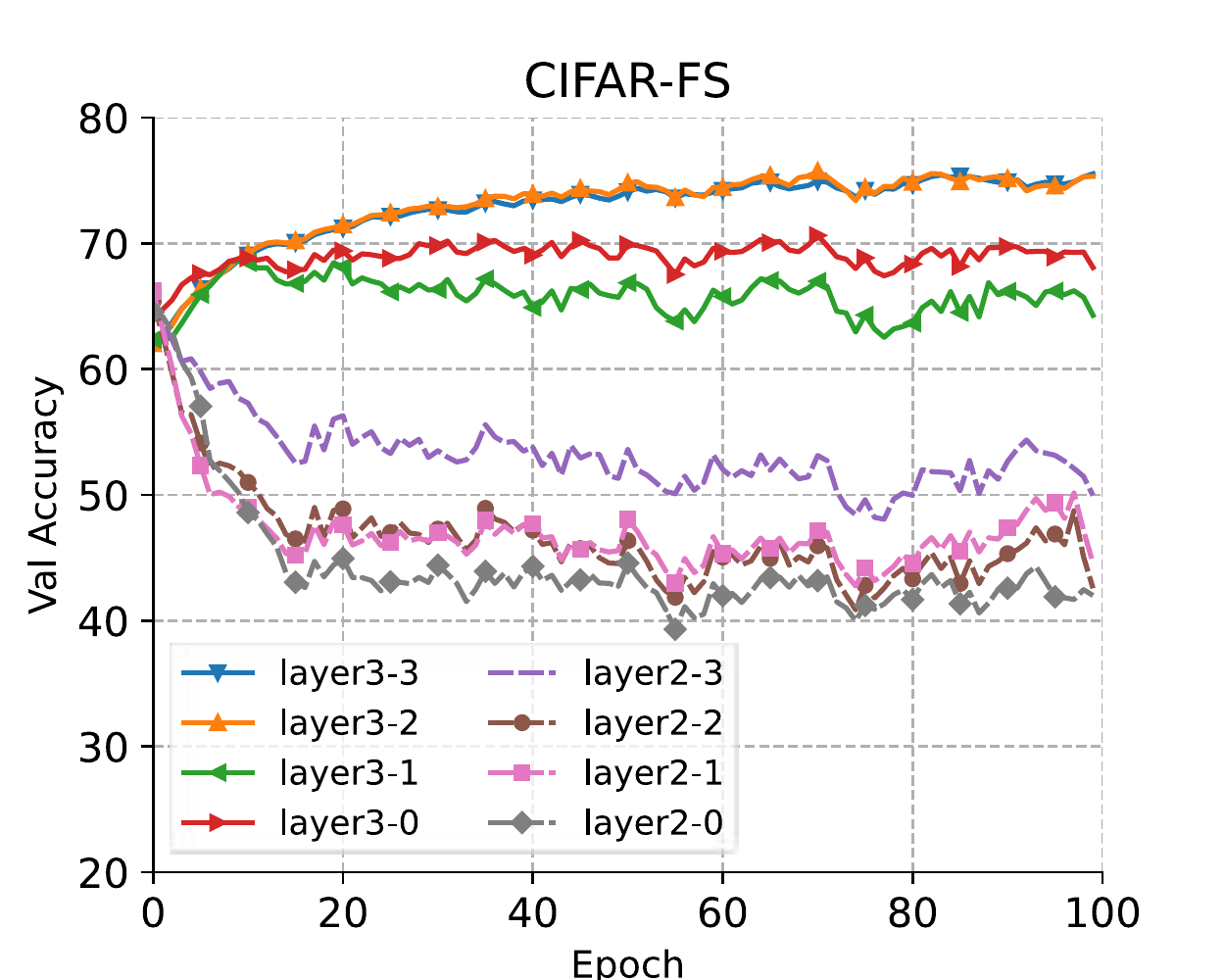}}
    \vspace{-0.3cm}
    \caption{Accuracy vs. different layers to inset prompts. We report 5-way 1-shot accuracy (\%) on the validation set of \textit{mini}ImageNet and CIFAF-FS along the meta-training process. The feature extractor has three stages and multiple Transformer layers in each stage.} 
    \label{fig:stage}
\vspace{-0.4cm}
\end{figure}

Theoretically, the semantic prompt in this work can be inserted into the feature extractor at any layer. However, we find that the layer selection has a significant impact on the performance. In Figure \ref{fig:stage}, we can see that inserting prompts at higher layers improves accuracies, while inserting prompts at lower layers leads to performance drop. Considering that prompt vectors are class-specific, these results indicate that class-specific features should be extracted at higher network layers, while features at lower layers should better be shared among classes. When looking into the performance of each layer, we can see that while the optimal layer selection varies slightly for different datasets, SP at all layers of the third stage improves accuracy consistently. To simplify architecture design, we choose the layer3-2 as default in our experiments.

\vspace{-0.1cm}
\subsubsection{The backbone and classifier architectures}

In Table \ref{tab:backbone}, we re-implement three baseline methods with the same Visformer backbone as ours, and compare the results with different backbones under the miniImageNet 1-shot setting. It can be seen that simply replacing ResNet-12 with Visformer can not obtain significant improvement. Instead, using semantic prompt can improves 1-shot performance over these baselines when equipped with the same Visformer backbone.

In Tab.\ref{tab:classifier}, we compare the LR and NN classifiers over all datasets. The simple NN classifier performs as well as the LR classifier for 1-shot, while the LR benefits from more training examples and outperforms the NN by 0.53\% for 5-shot.

\begin{table}[t]
    \footnotesize
    \centering
    \setlength{\tabcolsep}{0.6mm}{
    \begin{tabular}{lccccc}
        \midrule
        Backbone & ProtoNet\cite{snell2017prototypical} & MetaOptNet\cite{Lee2019MetaLearningWD} & Meta-Baseline\cite{chen2020new}  & Ours \\
        \midrule
        ResNet-12 & 63.28 & 63.29 & 64.36 & - \\
        Visformer-T & 63.16 & 64.39 & 63.32 & 72.31 \\
        \midrule
    \end{tabular}}
    \vspace{-0.3cm}
    \caption{Comparison with different backbones.}
    \label{tab:backbone}
    \vspace{-0.3cm}
\end{table}

\begin{table}[t]
    \footnotesize
    \centering
    \setlength{\tabcolsep}{1mm}{
    \begin{tabular}{ccccccccc}
        \midrule
        & \multicolumn{2}{c}{Mini} & \multicolumn{2}{c}{Tiered} & \multicolumn{2}{c}{CIFAR-FS} & \multicolumn{2}{c}{FC100}\\
        Classifier & 1-shot & 5-shot & 1-shot & 5-shot & 1-shot & 5-shot & 1-shot & 5-shot \\
        \midrule
        NN & 72.31 & 82.86 & 78.03 & 87.74 & 82.18 & 88.04 & 48.53 & 61.10 \\
        LR & 72.37 & 83.42 & 78.11 & 88.64 & 82.17 & 88.24 & 48.61 & 61.55 \\
        \midrule
    \end{tabular}
    \vspace{-0.3cm}
    \caption{Comparison of classifiers. \textbf{NN}: cosine-distance nearest prototype classifier. \textbf{LR}: linear logistic regression classifier.}
    \label{tab:classifier}}
\end{table}

\subsubsection{Projector structure and pooling strategy} 

\begin{table}
\small
\centering
    \begin{tabular}{lccccc}
    \toprule
     & \multicolumn{2}{c}{Projector} &  \multicolumn{3}{c}{Pooling strategy} \\
     \cmidrule(lr){2-3} \cmidrule(lr){4-6}
     & Linear & MLP & Head & Patches & All \\
    \midrule
    1-shot & 72.31 & 72.70 & 66.48 & 72.29 & 72.31 \\
    5-shot & 83.42 & 83.56 & 72.70 & 83.39 & 83.42 \\
    \bottomrule
    \end{tabular}
\caption{Choice of the projector, and the pooling strategy for the output sequence. `\textbf{Head}' means selecting the output at the position of the prompt vector; `\textbf{Patches}' means averaging the output features of all patches; `\textbf{All}' means averaging all feature vectors in the output sequence.}
\label{tab:architecture}
\vspace{-0.3cm}
\end{table}

As shown in Table \ref{tab:architecture}, the projector design has little effect on performance: both linear and MLP projectors work well and the MLP has slight advantage. In contrast, the pooling strategy has much more effect on performance. When adopting the `Head' strategy, both 1-shot and 5-shot learning accuracies are very poor. This indicates that the output at the position of the prompt vector is easy to overfit on semantic features and neglect rich visual features in image patches. Adopting average on all output features can address this problem and achieve better results.

\subsubsection{Image size and stem design}
\label{sec:image_size}

\begin{table}[t]
\small
\centering
    \begin{tabular}{lccc}
    \toprule
    Input size & 224$\times$224 & 84$\times$84 & 84$\times$84\\
    \midrule
    Stem & \makecell[c]{Ks=7,\\Stride=2} & \makecell[c]{Ks=7,\\Stride=2} & \makecell[c]{Ks=3,\\Stride=1} \\
    \midrule
    MiniImageNet & 72.31$\pm$0.40 & 68.09$\pm$0.38 & 72.16$\pm$0.40 \\
    TieredImageNet & 78.03$\pm$0.46 & 72.14$\pm$0.47 & 77.28$\pm$0.46 \\
    CIFAR-FS & 82.18$\pm$0.40 & 77.26$\pm$0.42 & 82.00$\pm$0.41 \\
    FC100 & 48.53$\pm$0.38 & 46.44$\pm$0.40 & 48.52$\pm$0.40 \\
    \bottomrule
    \end{tabular}
\vspace{-0.2cm}
\caption{The effect of input size and stem design. `Ks' means the kernel size of the first convolution layer (stem), and `Stride' means its stride. 5-way 1-shot accuracy is reported on four datasets with 95\% confidence intervals.}
\label{tab:image_size}
\end{table}

In Table \ref{tab:image_size}, we experiment with a smaller input size, 84$\times$84, to validate the influence of image size. It can be seen that directly changing the input size to 84$\times$84 leads to evident performance drop on all datasets. We suppose that this is because the kernel size and the stride of the stem is too large to capture the detailed visual features when the input image gets small. To address this problem, we reduce the kernel size and the stride of the stem accordingly. After this change, the 1-shot learning performance under 84$\times$84 improves significantly, and gets comparable results with the 224$\times$224 resolution on all datasets.

\subsubsection{Visualization}

\begin{figure}[t]
    \centering
    \includegraphics[width=\linewidth]{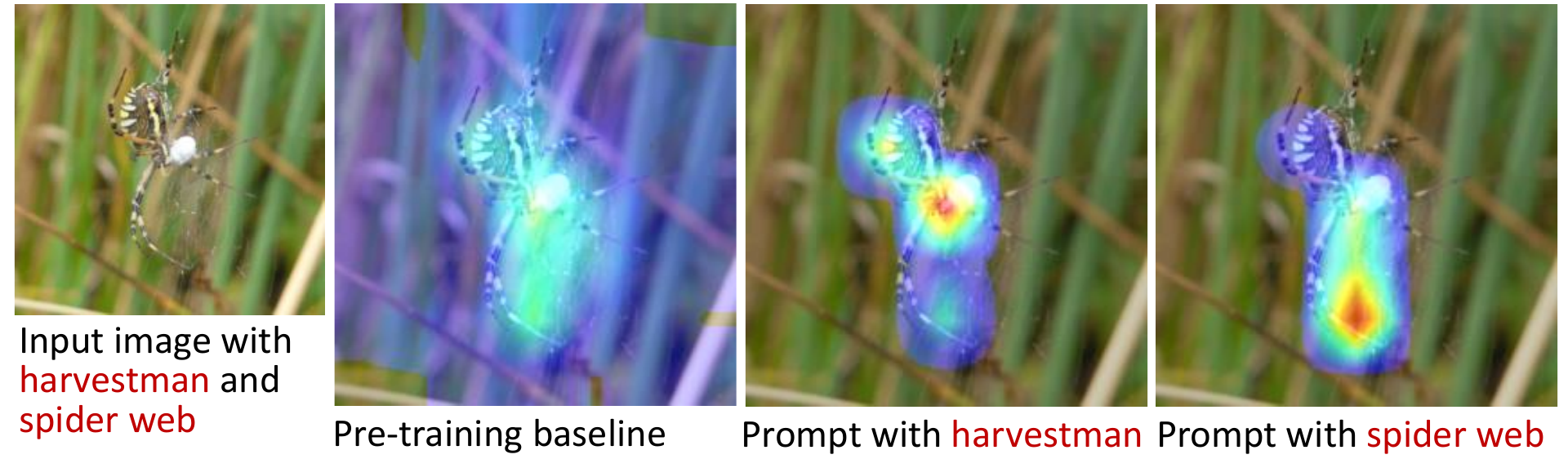}
    \caption{Visualization of attention maps when prompting with different class labels.} 
    \label{fig:visualization}
\vspace{-0.4cm}
\end{figure}

In Figure \ref{fig:visualization}, we visualize the attention maps by computing the dot product between the output feature and the feature vector at each location. It can be seen that the visual features of the pre-training baseline are cluttered with background information, but our method can focus on semantic-level visual features according to the given text prompt. For example, given the text prompt of harvestman, the model will attend to the features of the harvest rather than spider web or background clutters.

\section{Conclusion}

In this paper, we propose a novel Semantic Prompt (SP) approach for FSL, which adaptively tunes the feature extraction with the semantic features derived from class names. The proposed approach is evaluated on four benchmark datasets, and achieves significant improvements against previous methods. More in-depth analysis demonstrates that SP encourages the model to extract more class-specific features and is robust to different text encoders and model designs.

\section*{Acknowledgement}

This work was supported in part by the National Natural Science Foundation of China under Grants 61721004, 61976214, 62076078, 62176246 and National Key R\&D Program of China (2022ZD0117901).



{\small
\bibliographystyle{ieee_fullname}
\bibliography{main}
}

\end{document}